\documentclass[]{gewu}

\usepackage{wasysym}

\usepackage[toc,page,header]{appendix}

\usepackage{minitoc}
\usepackage{microtype}      
\usepackage[dvipsnames,table]{xcolor}         

\usepackage{hyperref}
\usepackage{url}
\usepackage{booktabs}
\usepackage{multirow}
\usepackage{subcaption}
\usepackage{caption}
\usepackage{amssymb}
\usepackage{graphicx}
\usepackage{amsmath}
\usepackage{wrapfig}
\usepackage{enumitem}
\usepackage{arydshln} 
\usepackage{gensymb}
\usepackage{makecell}
\usepackage{pifont} 
\usepackage{float}

\newcolumntype{B}{>{\columncolor{blue!2}}c}
\newcolumntype{O}{>{\columncolor{orange!4}}c}

\title{Robo-ValueRL: Reliable Value Estimation for Offline-to-Online Reinforcement Learning}

\author[1,*]{\small Wenke Xia}
\author[2,*]{\small Pei Ren}
\author[3]{\small Wenbo Yu}
\author[1]{\small Yizhuo Zhang}
\author[1]{\small Jifan Li}
\author[4]{\small Yixue Zhang}
\author[2]{\small Yinuo Zhao}
\author[2]{\small Qingyang Gao}
\author[5]{\small Jianlong Fu}
\author[2]{\small Jian Tang}
\author[1]{\small Ji-Rong Wen}
\author[2,\textnormal{\Letter}]{\small Zhengping Che}
\author[1,\textnormal{\Letter}]{\small Di Hu}

\affiliation[1]{Gaoling School of Artificial Intelligence, Renmin University of China}
\affiliationbreak
\affiliation[2]{Beijing Innovation Center of Humanoid Robotics}
\affiliationbreak
\affiliation[3]{Beijing Forestry University}
\affiliation[4]{Peking University}
\affiliation[5]{Microsoft Research }

\contribution[*]{Equal contribution}
\contribution[\textnormal{\Letter}]{ Corresponding author}

\abstract{
Offline-to-online reinforcement learning is promising for generalizable robotic manipulation, yet its full-stack complexity obscures reproduction and diagnosis. Within such systems, value estimation plays a central role in prioritizing heterogeneous data for policy improvement. Despite its importance, the central question remains underexplored: \textit{how value-function reliability shapes policy optimization in offline-to-online reinforcement learning}. To answer this question, we propose \textbf{Robo-ValueRL}, a unified framework that enables reliable value estimation and systematically traces its downstream effects on policy pretraining and online improvement. Concretely, Robo-ValueRL learns a history-conditioned value estimator and evaluates its reliability through global-progress and local-preference metrics. These resulting value estimates are propagated into quality-conditioned consistency-policy pretraining and a residual adaptation module on online rollouts, providing a unified testbed for analyzing how value reliability shapes downstream policy performance.
Across 240 hours of offline demonstrations and over 3,000 online rollout trajectories, our extensive experiments show that downstream performance is strongly associated with value reliability.
Reliable value functions provide better action-quality estimates, allowing \textbf{value-guided offline RL to scale more effectively than quality-agnostic behavior cloning}, and \textbf{stabilize online improvement by prioritizing high-quality rollout data}.
Integrating reliable value guidance through offline pretraining with online improvement, our system achieves \textbf{86\% success on millimeter-level precise chip insertion and 84\% on generalizable block disassembly}.
We hope these findings highlight the importance of value-guided data utilization for effective policy improvement from heterogeneous robotic experience.
}

\MyEmail{Wenke Xia at \email{xiawenke2022@ruc.edu.cn}, Di Hu at \email{dihu@ruc.edu.cn}, 
\emailbreak
Pei Ren at \email{pei.ren@x-humanoid.com}, Zhengping Che at \email{z.che@x-humanoid.com}}

\checkdata[Project Page]{\url{https://gewu-lab.github.io/Robo-ValueRL/}}
\checkdata[Code]{\url{https://github.com/Open-X-Humanoid/Robo-ValueRL}}
\checkdata[Data\&Model]{\url{https://huggingface.co/collections/X-Humanoid/robo-valuerl}}

\begin{document}
\maketitle


\vspace{-2em}
\section{Introduction}

Offline-to-online reinforcement learning has emerged as a promising paradigm for generalizable robotic manipulation~\cite{nair2020awac,kumar2020conservative,nakamoto2023cal}, bridging offline policy pretraining~\cite{brohan2022rt,zitkovich2023rt,o2024open,kim2024openvla,black2024pi_0,bjorck2025gr00t,kumar2022pre,zhang2026robogene} with online improvement through real-world interaction~\cite{luo2025precise,yang2024robot,li2025simplevla,xia2025phoenix,NEURIPS2025_347110fb,Yu_2026_CVPR}. However, unlike conventional benchmarks, it constitutes a full-stack embodied learning system shaped by coupled components, including \textit{data curation}, \textit{value estimation}, \textit{offline pretraining}, and \textit{online exploration}~\cite{henderson2018deep,agarwal2021deep,wang2026learning}. These interdependencies make performance difficult to reproduce and failures hard to diagnose under deployment conditions.

Among these factors, value estimation is central to offline-to-online robotic reinforcement learning~\cite{feng2024suf,lee2022offline}, as it evaluates heterogeneous experience, from suboptimal demonstrations to online rollouts, and determines how such data contribute to policy improvement~\cite{intelligence2025pi,mandlekar2021matters,kalashnikov2018scalable}. Reliable values can prioritize informative experience and suppress low-quality data, whereas inaccurate values may misrank experience and destabilize learning. However, existing evidence on value-estimation reliability remains fragmented across the offline-to-online pipeline~\cite{nakamoto2023cal,intelligence2025pi,li2025gr,ma2022vip}: algorithmic studies often analyze value optimization in isolation~\cite{kumar2020conservative,nakamoto2023cal}, while robotic systems mainly report downstream performance without examining whether learned values capture task progress or action quality~\cite{intelligence2025pi,li2025gr,chen2025conrft}. Thus, \textit{the role of value-estimation reliability in offline pretraining and online improvement has yet to be systematically investigated.}

\begin{figure*}[t]
    \centering
\includegraphics[width=0.9\columnwidth]{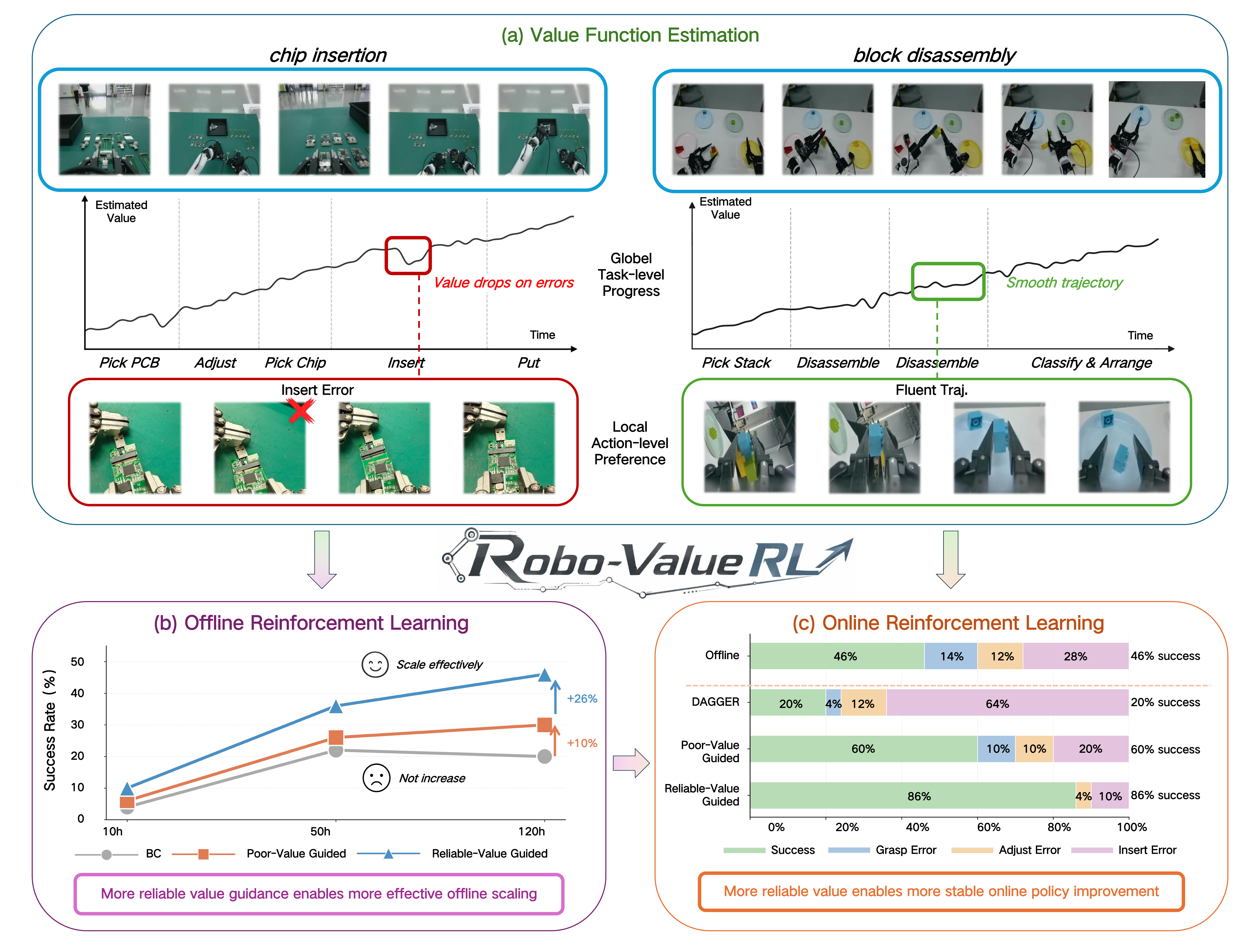}
\caption{
Robo-ValueRL improves offline-to-online RL via reliable value guidance. 
(a) Value functions are assessed with reliability metrics for downstream policy learning.
(b) Reliable values enable scalable offline pretraining from mixed-quality demonstrations. 
(c) Value-guided online adaptation stabilizes policy improvement, reaching 86\% millimeter-level chip-insertion success.
}
    \vspace{-1em}
    \label{fig:teaser}
    
\end{figure*}

To investigate this role, we examine value-estimation reliability from two coupled perspectives: how reliable values are learned from heterogeneous robotic data, and how their reliability affects policy learning across offline pretraining and online improvement. To ground this investigation, we introduce Robo-ValueRL, a unified offline-to-online robotic reinforcement learning framework that organizes \textit{reliable value function learning}, \textit{value-guided offline policy pretraining}, and \textit{real-world online improvement} into a coherent process. Beyond final success rates, Robo-ValueRL introduces value-reliability metrics that provide an analytical basis for relating value quality to downstream policy optimization.


Concretely, Robo-ValueRL first trains a history-conditioned value estimator on offline robotic data to reduce ambiguity from partial observations and visual occlusions. Its action-quality estimates are used to score offline experience for quality-conditioned consistency-policy pretraining~\cite{prasad2024consistency}, enabling efficient value-guided action generation through one-step denoising. The pretrained policy is then deployed to collect real-world rollouts with human-in-the-loop intervention. Robo-ValueRL further updates the value estimator on these rollouts and selects high-quality online segments to train a lightweight residual adaptation module, allowing targeted correction of failure modes while preserving prior knowledge. Across this offline-to-online process, our metrics assess value reliability from both global task-level progress to measure whether values reflect long-horizon task advancement, and local action-level preference to estimate whether values capture fine-grained action quality. Together, Robo-ValueRL provides a unified testbed for analyzing how value reliability affects offline scaling and online improvement.

Leveraging Robo-ValueRL, we conduct an extensive real-world study on two challenging manipulation tasks: millimeter-level chip insertion and generalizable block disassembly, using 240 hours of offline robotic data and over 3,000 online rollout trajectories. As summarized in Figure~\ref{fig:teaser}, our results reveal a consistent relationship between downstream policy performance and value-estimation reliability: value functions that capture both \textit{global task progress} and \textit{local action preference} enable more effective utilization of heterogeneous robotic experience.
During offline pretraining, \textbf{value-guided policy learning scales more effectively than quality-agnostic behavior cloning} under mixed-quality demonstrations, improving offline success rates by 26\% on chip insertion and 34\% on block disassembly. During online improvement, \textbf{reliable value estimates lead to more stable real-world adaptation} by prioritizing high-quality rollout segments and suppressing suboptimal interactions. Overall, Robo-ValueRL achieves \textbf{86\% success on chip insertion and 84\% on block disassembly}, demonstrating that reliable value-guided data utilization is critical for converting heterogeneous offline and online experience into effective robotic policy improvement.

\section{Related Work}

\subsection{Learning from Heterogeneous Robotic Data}

A central challenge in scaling robotic learning is the heterogeneity of real robot data~\cite{cao2021learning,chen2025curating}: human demonstrations contain mistakes, hesitation, and locally suboptimal actions, while online rollouts introduce exploratory failures and unstable interactions~\cite{wang2026learning,chen2025conrft}. Current large-scale VLA pipelines typically absorb such mixed-quality data through behavior cloning, which provides broad behavioral priors but lacks explicit action-quality discrimination~\cite{black2024pi_0,team2024octo,fan2025xr}. Consequently, quality-agnostic imitation may fit both effective and suboptimal segments. Offline RL and recent offline-to-online methods address this limitation by using value functions to exploit higher-quality behaviors from offline data and online rollouts~\cite{prudencio2023survey,kostrikov2021offline,nakamoto2023cal,intelligence2025pi,li2025gr}. However, existing work mainly evaluates heterogeneous data usage through final policy performance, leaving unclear how value estimation is reliable and how such reliability affects downstream learning.
We address this gap by treating value estimation as an interface between heterogeneous robotic data and policy learning, enabling mixed-quality trajectories to be incorporated into policy optimization.


\subsection{Offline-to-Online Reinforcement Learning}
Offline-to-online reinforcement learning combines the sample efficiency of offline RL with the adaptability of online interaction~\cite{nair2020awac,zhang2023policy}. 
A key challenge is distribution shift: once deployed, an offline-trained policy may visit poorly covered states and actions, causing critic miscalibration and unstable online updates~\cite{chebotar2023q,luo2023finetuning}. 
Existing methods mitigate this issue by constraining online adaptation around the offline policy and balancing offline-online data usage to preserve useful priors while adapting safely~\cite{nakamoto2023cal,lee2022offline}. 
However, these studies are often evaluated on small-scale benchmarks. 
Recent VLA-based robotic systems~\cite{intelligence2025pi,li2025gr, chen2025conrft} extend offline-to-online learning to large generalist policies, but mainly report final performance, leaving the stability and design dependencies of value estimation underexplored. 
In this work, we build the Robo-ValueRL framework to systematically study how value estimation affects offline policy pretraining and online improvement, and whether value-function reliability can be diagnosed before downstream policy learning.

\subsection{Value Estimation in Robotic Manipulation}

Value estimation is central to robotic manipulation, where critics provide long-horizon signals for policy improvement. From temporal-difference learning and linear approximation to deep RL for visual control~\cite{sutton1988learning,tsitsiklis1996analysis,sutton2009fast,kalashnikov2018scalable,mnih2015human,duan2016benchmarking,haarnoja2018soft}, value functions have traditionally been learned as policy-coupled critics. Recent work instead trains general value functions to evaluate task progress and behavior quality across goals, tasks, and datasets~\cite{zhai2025vision,wang2024rl,yang2024rank2reward,ma2025vision}, with goal-conditioned values, successor representations, and world-value models further improving transferable value prediction~\cite{ma2022vip,schaul2015universal,lv2026viva,wang2026world}. However, evaluation has not kept pace: existing metrics such as Value-Order Correlation in GVL and smoothness in VIP mainly assess optimal trajectories~\cite{ma2025vision,ma2022vip}, while concurrent work studies diverse suboptimal value estimation but validates it primarily through filtered behavior cloning~\cite{wang2026world}. We instead train a history-conditioned value estimator and introduce global task-level progress and local action-level preference metrics to assess its role across offline-to-online learning.

\begin{figure*}[t]
    \centering
\includegraphics[width=\columnwidth]{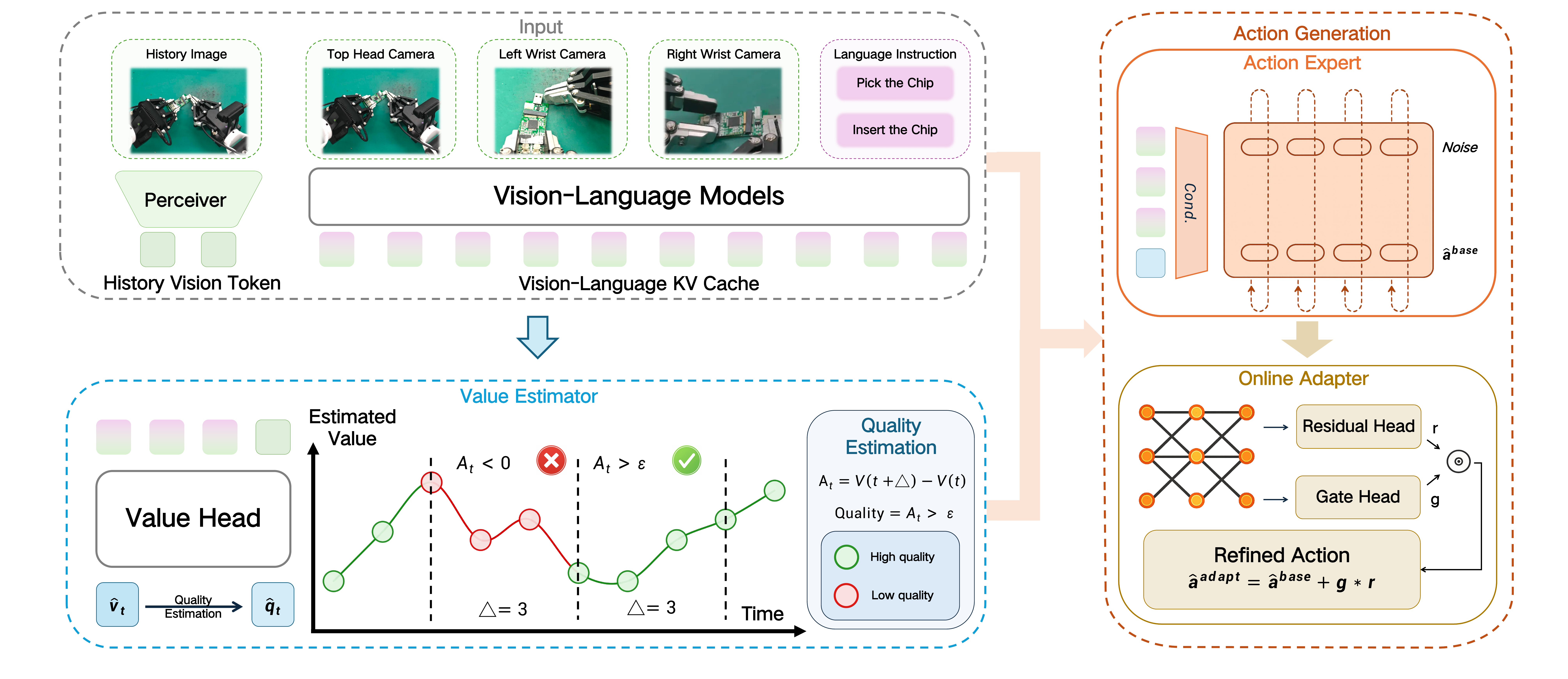}
    \caption{Our Robo-ValueRL offline-to-online reinforcement learning framework. The value estimator incorporates historical context to produce reliable value estimates, from which action-quality labels are derived to train a quality-conditioned consistency policy.
    During online interactions, the online adaptation module is further utilized to enable stable online improvement. 
    }
    \label{fig:pipeline}
    \vspace{-1em}
\end{figure*}

\section{Robo-ValueRL}

We propose Robo-ValueRL, a unified offline-to-online robotic reinforcement learning framework centered on reliable value estimation. Robo-ValueRL first trains a history-conditioned value estimator to infer normalized task progress from visual observations and history, and uses value differences to derive action-quality indicators for quality-conditioned VLA consistency-policy pretraining. For online improvement, it freezes the pretrained policy and learns a lightweight residual adapter from real-world rollouts, enabling targeted adaptation while preserving the offline prior. We further introduce value-reliability metrics that assess global task-stage ordering and local action-level preference, linking value quality to both offline scaling and online policy improvement.

\subsection{History-Conditioned Value Estimator}\label{subsec:value_estimator}

Single-frame value estimation is inherently ambiguous in long-horizon robotic manipulation, where occlusions and repetitive visual patterns can make different task stages appear visually similar. To address this, we introduce a history-conditioned value estimator that augments the current visual observation with a compact representation of past observations for more reliable value prediction, with implementation details provided in Appendix~\ref{sec:value_estimator_sup}.

The value estimator uses a PaliGemma vision-language backbone~\cite{beyer2024paligemma} with a lightweight Transformer value head $f_\theta^v$. At timestep $t$, the current three-view observation $o_t$ and language instruction $\ell$ are processed by PaliGemma to produce a vision-language conditioning context $\mathbf{c}_t^{v}$, which serves as the VLM-generated prefix representation for subsequent value estimation. Further, the visual history $h_t$ is encoded via a SigLIP encoder~\cite{zhai2023sigmoid} and summarized into compact tokens $\tilde{\mathbf{H}}_t$ using a lightweight Perceiver module~\cite{jaegle2021perceiver}.
To predict the task value, we append a learnable value query $\mathbf{q}_v$ to $\tilde{\mathbf{H}}_t$ and employ a transformer-based value head conditioned on $\mathbf{c}_t^{v}$. The network outputs a categorical distribution over $K$ discretized value bins $\{b_k\}_{k=1}^{K}$:
\begin{equation}
    p_\theta(k \mid o_t, h_t, \ell) = \mathrm{softmax}_k \left( f_\theta^v \left( \mathbf{c}_t^{v}, \tilde{\mathbf{H}}_t, \mathbf{q}_v \right) \right).
\end{equation}

\textbf{Training Recipe.}
We supervise estimator using a normalized progress value $v_t^* \in [0, 1]$. We adopt an HL-Gaussian distributional representation~\cite{li2025gr,farebrother2024stop}, the target $v_t^*$ is projected onto the bins as a soft distribution $q_k(v_t^*) \propto \exp\left({-\frac{(v_t^* - b_k)^2}{2\sigma^2}}\right)$. The value head is optimized via cross-entropy loss:
\begin{equation}
    \mathcal{L}_{v} = - \sum_{k=1}^{K} q_k(v_t^*) \log p_\theta(k \mid o_t, h_t, \ell).
\end{equation}
During inference, the final scalar value estimate is decoded as the expectation over the bins: $\hat{v}_t = V_\theta(o_t,h_t,\ell) = \sum_{k=1}^{K} b_k \cdot p_\theta(k \mid o_t, h_t, \ell)$.

\textbf{Action Quality Indicator.}
The value estimator provides dense action-quality indicators to guide both offline reinforcement learning and online fine-tuning. Intuitively, an action is deemed beneficial if it increases the estimated task progress. Formally, for a transition from $t$ to $t+\Delta$, we compute the value improvement as $\Delta v_t = V_\theta(o_{t+\Delta}, h_{t+\Delta}, \ell) - V_\theta(o_t, h_t, \ell)$. We then assign a discrete action quality indicator $q_t^{\mathrm{act}} \in \{0, 1, 2\}$ (representing low-quality, neutral, and high-quality actions, respectively) by thresholding $\Delta v_t$ with pre-defined positive and negative margins $\delta_{+}$ and $\delta_{-}$.

\subsection{Quality-conditioned Vision-Language-Action Model}\label{subsec:policy}

Given the action-quality indicators estimated from offline experience, we train a quality-conditioned VLA model $\pi_{\theta}$ with a consistency policy head $f_{\theta}^a$ on heterogeneous demonstrations. The policy learns to associate action patterns with estimated quality, separating high-quality behaviors from suboptimal segments. During inference, the desired quality prompt guides the policy toward high-quality action generation, while the consistency head enables one-step denoising for efficient real-time control. Additional implementation details are provided in Appendix~\ref{sec:policy_sup}.

Concretely, our quality-conditioned VLA model $\pi_{\theta}$ uses PaliGemma as the vision-language backbone and a consistency policy head $f_\theta^a$ as action expert. At timestep $t$, to inject value-based quality information into the language-conditioned policy, we first convert the action-quality indicator $q_t^{\mathrm{act}}$ into a textual action-quality prompt $\ell_t^{\mathrm{act}}$ and concatenate it with the task instruction to form $\ell_t^{\mathrm{qc}} = [\ell; \ell_t^{\mathrm{act}}]$. Further, the VLM model takes the multi-view observation $o_t$ and quality-conditioned language input $\ell_t^{\mathrm{qc}}$ as input to generate vision-language context $\mathbf{c}_t^{a}$ for action generation.

To enable efficient action generation, we instantiate action expert $f_{\theta}^a$ as a consistency policy~\cite{prasad2024consistency}. Conditioned on the VLM-generated action context $\mathbf{c}_t^{a}$ and the robot state $s_t$, the policy maps a noisy action chunk $\mathbf{x}_{\sigma_i}$ at noise level $\sigma$ directly to the corresponding clean action chunk: $\hat{\mathbf{x}}_0 = f_{\theta}^a(\mathbf{x}_{\sigma_i}, \sigma_i, s_t, \mathbf{c}_t^{a})$.
During inference, we sample $\mathbf{x}_{\sigma_{\max}} \sim \mathcal{N}(0, \sigma_{\max}^2 I)$ and obtain the final action through a single denoising step, enabling low-latency real-time robot execution.


\textbf{Training Recipe.}
The quality-conditioned VLA model $\pi_{\theta}$ is optimized with a joint objective that combines action reconstruction and adjacent-noise self-consistency. Given a ground-truth action chunk $\mathbf{x}_0$, we sample a noise level $\sigma_i$ from a Karras noise schedule to construct a noisy action chunk: $\mathbf{x}_{\sigma_i}=\mathbf{x}_0 + \sigma_i \boldsymbol{\epsilon}$, where $\boldsymbol{\epsilon} \sim \mathcal{N}(0,I)$. 
The model is first supervised by an action reconstruction loss:
\begin{equation}
\mathcal{L}_{\mathrm{act}} = \| \pi_\theta(\mathbf{x}_{\sigma_i}, \sigma_i, s_t, o_t, \ell_t^{\mathrm{qc}}) - \mathbf{x}_0 \|_2^2.
\end{equation}
We further impose self-consistency between adjacent noise levels $\sigma_i$ and $\sigma_{i+1}$ to align clean-action predictions along the denoising trajectory. The prediction at $\sigma_i$ is propagated by a one-step solver to $\tilde{\mathbf{x}}_{\sigma_{i+1}}$, yielding the consistency loss:
\begin{equation}
\mathcal{L}_{\mathrm{cons}} = w(\sigma_i) \| \pi_\theta(\mathbf{x}_{\sigma_i}, \sigma_i, s_t, o_t, \ell_t^{\mathrm{qc}}) - \mathrm{sg}[ \pi_{\bar{\theta}} ( \tilde{\mathbf{x}}_{\sigma_{i+1}}, \sigma_{i+1}, s_t, o_t, \ell_t^{\mathrm{qc}} ) ] \|_2^2,
\end{equation}
where $w(\sigma_i)$ is a noise-dependent weighting term, $\mathrm{sg}[\cdot]$ denotes the stop-gradient operation, and $\pi_{\bar{\theta}}$ is updated as an exponential moving average of the online model $\pi_{\theta}$. This consistency regularization encourages the full VLA model to produce stable clean-action predictions across adjacent noise levels, enabling the learned policy to approximate multi-step denoising with a single denoising step during inference. 
The final joint training objective is balanced by a hyperparameter $\lambda_{\mathrm{cons}}$:
\begin{equation}
    \mathcal{L} = \mathcal{L}_{\mathrm{act}} + \lambda_{\mathrm{cons}}\mathcal{L}_{\mathrm{cons}}.
\end{equation}


\subsection{Online Residual Policy Adaptation}\label{subsec:adapter}
Online rollouts provide valuable interaction data for policy improvement, but directly fine-tuning the pretrained VLA policy on heterogeneous rollouts can destabilize optimization and induce catastrophic forgetting of the learned behavior prior. We therefore freeze the offline-trained base policy $\pi_{\theta}^{\mathrm{base}}$ and introduce a lightweight residual adapter $h_{\phi}$, which predicts bounded corrections to the base action to enable targeted adaptation from online experience while preserving the pretrained prior. Additional architectural and training details are provided in Appendix~\ref{sec:adapter_module_sup}.

Given the action chunk $\hat{\mathbf{x}}_{t}^{\mathrm{base}}$ from the frozen base policy $\pi_{\theta}^{\mathrm{base}}$, the residual adapter $h_{\phi}$ predicts a bounded correction to refine it. Specifically, the adapter compresses the VLA action context $\mathbf{c}_t^{a}$ into compact latent context tokens using a Perceiver module~\cite{jaegle2021perceiver}, concatenates them with the robot state $s_t$ and base action prediction $\hat{\mathbf{x}}_{t}^{\mathrm{base}}$, and feeds the result into a lightweight Transformer to predict an unconstrained residual $\tilde{\mathbf{r}}_t$ and a step-wise gate logit $\tilde{\mathbf{g}}_t$:
\begin{equation}
\left(
\tilde{\mathbf{r}}_t,
\tilde{\mathbf{g}}_t
\right)
=
h_{\phi}
\left(
\mathbf{c}_t^{a},
s_t,
\hat{\mathbf{x}}_{t}^{\mathrm{base}}
\right),
\end{equation}
where we bound the residual as $\mathbf{r}_t = r_{\max}\tanh(\tilde{\mathbf{r}}_t)$ and constrain the gate with a sigmoid function as $\mathbf{g}_t = \sigma(\tilde{\mathbf{g}}_t)$,
The final adapted action chunk is computed as:
\begin{equation}
\hat{\mathbf{x}}_t^{\mathrm{adapt}}
=
\hat{\mathbf{x}}_t^{\mathrm{base}}
+
\mathbf{g_t}\odot\mathbf{r_t}.
\end{equation}

\textbf{Training Recipe.}
We train only the residual adapter $h_{\phi}$ with a mixture of offline demonstrations $\mathcal{D}_{\mathrm{off}}$ and online rollouts $\mathcal{D}_{\mathrm{on}}$, while keeping base VLA policy $\pi_{\theta}^{\mathrm{base}}$ frozen. We first use the learned value estimator to split online rollouts into high-quality samples $\mathcal{D}_{\mathrm{on}}^{+}$ and low-quality samples $\mathcal{D}_{\mathrm{on}}^{-}$ according to the estimated action-quality indicator $q_t^{\mathrm{act}}$. For high-quality online samples from $\mathcal{D}_{\mathrm{on}}^{+}$, the correction loss $\mathcal{L}_{\mathrm{corr}}=\mathbb{E}_{\mathcal{D}_{\mathrm{on}}^{+}}\left[\left|\left|\hat{\mathbf{x}}_{t}^{\mathrm{adapt}}-\mathbf{x}_{t}\right|\right|_2^2\right]$ encourages the adapted action to move toward successful online behaviors. For offline samples from $\mathcal{D}_{\mathrm{off}}$, the behavior-preservation loss $\mathcal{L}_{\mathrm{keep}}=\mathbb{E}_{\mathcal{D}_{\mathrm{off}}}\left[\left|\left|\hat{\mathbf{x}}_{t}^{\mathrm{adapt}}-\hat{\mathbf{x}}^{base}_{t}\right|\right|_2^2\right]$keeps the adapted action close to the frozen base action $\hat{\mathbf{x}}_{t}^{\mathrm{base}}$, preventing the residual adapter from drifting away from the pretrained behavior prior.
We further introduce a gate loss $\mathcal{L}_{\mathrm{gate}}=\mathbb{E}_{\mathcal{D}_{\mathrm{on}}^{+}\cup\mathcal{D}_{\mathrm{off}}}\!\left[\mathrm{CE}\!\left(\mathbf{g}_t,y_t^{g}\right)\right]$ to regulate when the residual correction should be activated. The final adapter objective is:
\begin{equation}
\mathcal{L}_{\mathrm{adapter}}
=
\mathcal{L}_{\mathrm{corr}}
+
\lambda_{\mathrm{keep}}
\mathcal{L}_{\mathrm{keep}}
+
\lambda_{\mathrm{gate}}
\mathcal{L}_{\mathrm{gate}}.
\end{equation}

\subsection{Value-reliability Metrics}~\label{subsec:metrics}

To systematically diagnose value-function reliability, we establish an evaluation protocol that comprehensively evaluates learned value functions across global task-level progress and local action-level preference. Given a predicted value sequence $\{\hat{v}_t\}$,  this suite verifies the model's capacity to resolve task advancement, identify high-quality actions, and detect execution errors.

\textbf{Global Task-level Progress.}
This diagnostic measures whether the value function preserves the coarse ordering of task progress. 
Given subgoal boundary timesteps $\{b_0,\dots,b_K\}$, we extract the midpoint of each interval as 
$m_j=\hat{v}_{\lfloor(b_j+b_{j+1})/2\rfloor}$. 
Since later intervals correspond to more advanced task stages, a reliable value function should assign higher values to later subgoals, i.e., $m_{j+1}>m_j$. 
We define the \emph{Midpoint Ordering Rate} (MOR) as the fraction of adjacent subgoal transitions that satisfy this ordering:
$\mathrm{MOR}=\frac{1}{K-1}\sum_{j=0}^{K-2}\mathbb{1}[m_{j+1}>m_j]$.

\textbf{Local Action-level Preference.}
This diagnostic evaluates whether the value function provides stable and accurate action-level estimates along local trajectories. A reliable value function should remain smooth during successful executions while sharply penalizing anomalous deviations from steady task progress. We characterize this behavior using two complementary metrics:

\textit{1) Fluency:} Along successful execution paths, the predicted value sequence should increase smoothly, whereas abrupt downward jumps indicate noisy or unstable value estimates. To quantify this local fluency, we evaluate both the frequency and the severity of these temporal oscillations~\cite{ma2022vip}. Specifically, we define the \textit{Bump Ratio} as $\frac{1}{T-1} \sum_{t} \mathbb{1}[\hat{v}_{t+1} < \hat{v}_t - \epsilon_v]$, which measures how often the expected upward trend is violated beyond a small noise-tolerance threshold $\epsilon_v$. Concurrently, we compute the \textit{Bump Magnitude} as $\frac{1}{T-1} \sum_{t} \max(0, \hat{v}_t - \hat{v}_{t+1})$ to capture the average magnitude of these downward drops, to quantify the overall stability of the value signals.

\textit{2) Temporal Error Discrimination:}
While fluency measures stability along successful executions, temporal error discrimination evaluates whether the value function can reliably penalize deviations from normal task progress, thereby preventing erroneous behaviors from being assigned overly optimistic action-quality estimates.
For each annotated error segment $\mathcal{E}$, we construct a normal-progress reference $\hat{v}^{\mathrm{ref}}_i$ from the pre-error value trajectory, representing the value that would be expected under uninterrupted task progress. 
We compare this reference with the smoothed prediction $\tilde{v}_i$ and define the value deficit as $\delta_i=\hat{v}^{\mathrm{ref}}_i-\tilde{v}_i$, where $\delta_i>0$ means that the estimator assigns a lower value than expected under normal progress. 
We report \textit{Error Sensitivity}, $S=\max_{i\in\mathcal{E}}\delta_i$, to measure the peak value deficit, and \textit{Error Slope}, $\alpha=\mathrm{slope}(\{\delta_i\}_{i\in\mathcal{E}})$, to measure whether this deficit is sustained over the erroneous execution.

\section{Experiments}

\begin{figure*}[t]
    \centering
\includegraphics[width=\columnwidth]{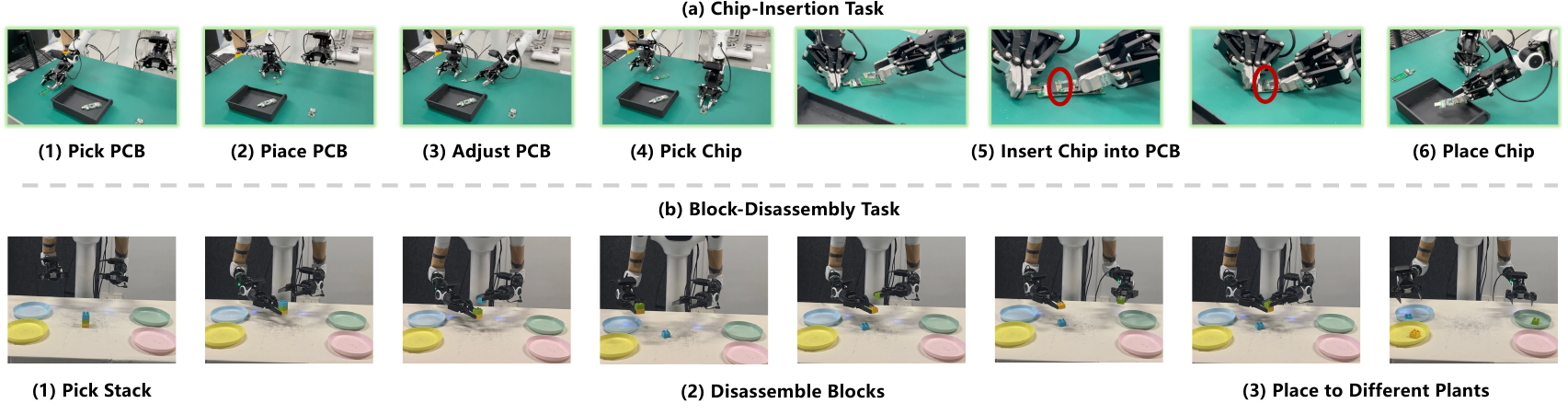}
    \caption{Overview of the two real-world manipulation tasks. (a) \textbf{Chip insertion}: a long-horizon, fine-grained task where the robot grasps a PCB, adjusts it to a feasible pose, and inserts a chip into a millimeter-scale slot within the recessed groove beneath the PCB base plate, as highlighted by the red circle. (b) \textbf{Block disassembly}: the robot disassembles randomly placed blocks and sorts them onto color-matched plates with randomized layouts, requiring generalizable bimanual coordination.}
    \label{fig:task}
\end{figure*}

In this work, we investigate \textit{how reliable values can be learned from heterogeneous robotic data} and \textit{how value reliability affects policy learning across offline pretraining and online improvement}. To this end, we build Robo-ValueRL, a unified framework for reliable value estimation that systematically traces its downstream effects on policy pretraining and online improvement. We evaluate Robo-ValueRL on two challenging real-world manipulation tasks with 240 hours of heterogeneous offline demonstrations and over 3,000 online rollout trajectories, as shown in Figure~\ref{fig:task}, covering both precision-critical fine-grained manipulation and generalizable bimanual coordination. Building on this experimental setting, we organize our investigation around three progressively connected questions:

\textbf{(1) How can value reliability be diagnosed?} In Section~\ref{sec:value_diagnostics}, we evaluate value-reliability metrics and examine their alignment with downstream policy success.

\textbf{(2) How does value reliability affect offline pretraining?} In Section~\ref{sec:offline_policy_learning}, we compare value-guided pretraining across data scales and quality thresholds, showing how reliable values improve learning from heterogeneous demonstrations.

\textbf{(3) How does value reliability affect online improvement?} In Section~\ref{sec:online_adaptation}, we study iterative online adaptation with value-guided rollout filtering, showing that reliable values stabilize policy improvement from real-world interaction.

Finally, in Section~\ref{sec:qualitative}, we provide qualitative evidence that complements the quantitative results, showing that robots learn efficient execution and autonomous self-correction patterns through our Robo-ValueRL offline-to-online reinforcement learning framework.

\subsection{Value Reliability and Its Downstream Effects}
\label{sec:value_diagnostics}

The lack of direct quantitative measures for value-function reliability makes it difficult to diagnose how value-estimation quality affects downstream policy optimization. We introduce value-reliability metrics to evaluate value functions and examine their alignment with downstream policy learning, and provide more visualization results in Appendix~\ref{sec:value_sup}.

\begin{figure*}[!t]
    \centering
    \vspace{-1em}
    \captionof{table}{
    Ablation study on value estimation diagnostics and downstream policy learning.
    \textsc{Short History} and \textsc{Long History} denote 5-frame and 30-frame past visual histories. MOR is the Midpoint Ordering Rate for global task-level progress estimation.}
    \label{tab:value_ablation}

    \vspace{-2mm}
    \setlength{\tabcolsep}{6pt}
    \renewcommand{\arraystretch}{1.15}
    \resizebox{\textwidth}{!}{
    \begin{tabular}{lBBBBBOO}
        \toprule
        \textbf{Variant}
        & \multicolumn{5}{c}{\cellcolor{blue!10}\textbf{Value Estimation Metrics}}
        & \multicolumn{2}{c}{\cellcolor{orange!15}\makecell[c]{\textbf{Downstream Success Rate}}} \\
        \cmidrule(lr){2-6}
        \cmidrule(lr){7-8}
        & \textbf{MOR. $\uparrow$}
        & \textbf{Bump Ratio $\downarrow$}
        & \textbf{Bump Mag. $\downarrow$}
        & \textbf{Sensitivity $\uparrow$}
        & \textbf{Slope $\uparrow$}
        & \textbf{Chip $\uparrow$}
        & \textbf{Block $\uparrow$} \\
        \midrule
        \textsc{No History}    & 95.2 & 0.073 & 0.74 & 35.2 & 0.54 & 28.0\% & 44.0\% \\
        \textsc{Short History} & \textbf{95.6} & \textbf{0.067} & 0.70 & \textbf{41.1} & \textbf{0.62} & \textbf{46.0\%} & \textbf{60.0\%} \\
        \textsc{Long History}  & 95.1 & 0.069 & \textbf{0.66} & 35.8 & 0.58 & 30.0\% & 42.0\% \\
        \bottomrule
    \end{tabular}
    }

    \vspace{2mm}

    \includegraphics[width=\textwidth]{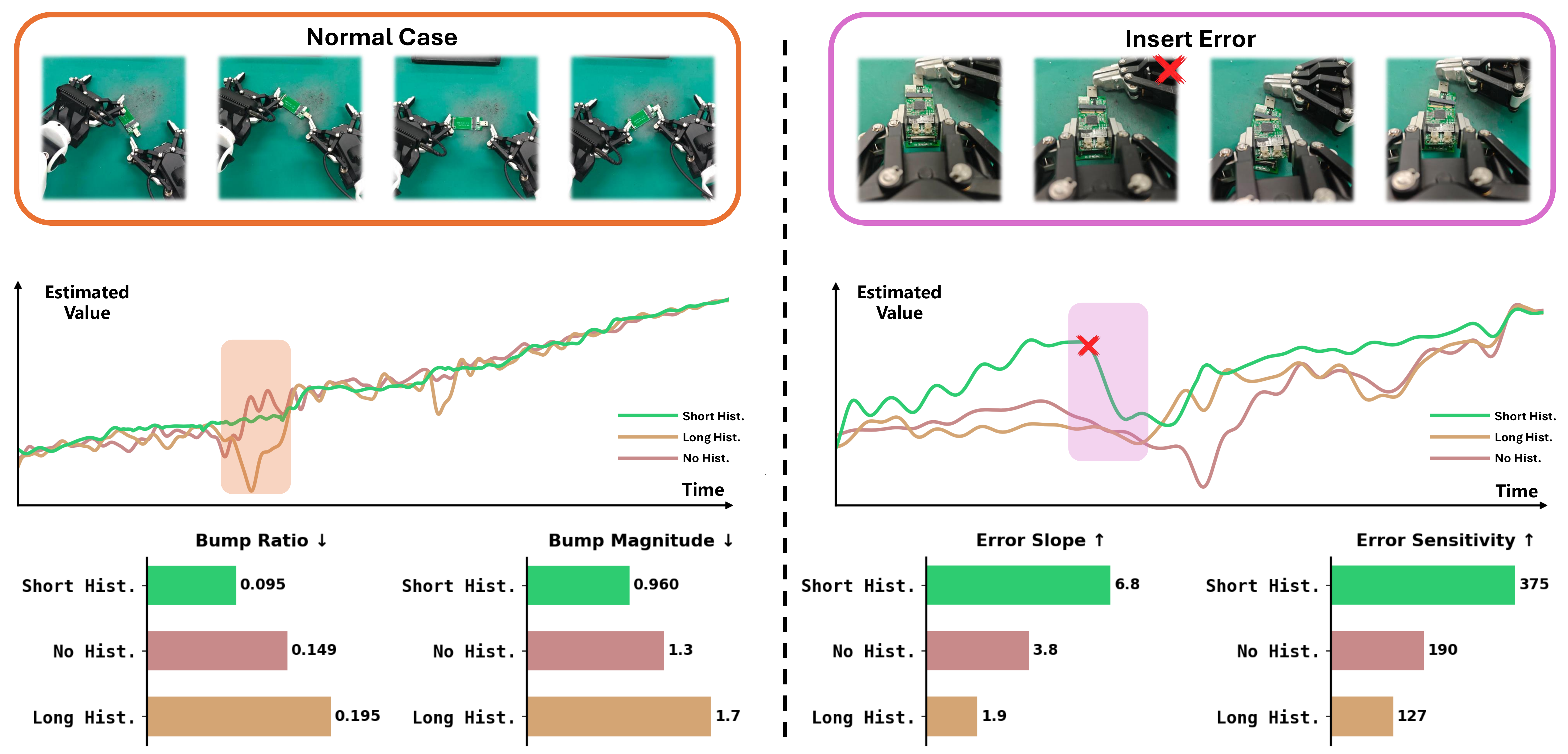}

    \vspace{-1mm}
    \captionof{figure}{
Estimated value curves with metrics. The metrics reflect stability in smooth trajectories and capture drops at error onset, demonstrating their effectiveness for measuring value reliability.
}
    \label{fig:value}
    \vspace{-1em}

\end{figure*}


\textbf{Reliability Metrics for Value Estimation.} We ablate the temporal context used by the value estimator: \textsc{No History} uses only the current observation, while \textsc{Short History} and \textsc{Long History} use 5-frame and 30-frame visual histories, respectively. The metrics columns in Table~\ref{tab:value_ablation} show that \textsc{Short History} provides the strongest overall value reliability. These results indicate that moderate temporal context helps preserve global progress ordering, suppress local value fluctuations, and produce clearer responses to erroneous actions. 
Figure~\ref{fig:value} visualizes the behaviors captured by these metrics. \textsc{Short History} produces smoother predictions during normal execution and a sharper drop after errors, corresponding to its stronger local fluency and error-response metrics. In contrast, other variants exhibit more oscillatory traces or weaker responses to erroneous actions, which are also reflected by their lower metrics scores. These traces support that our metrics quantitatively capture key properties of value reliability, including normal-trajectory stability and error sensitivity.


\textbf{Diagnostic Validity for Policy Learning.}
We further examine whether the proposed metrics predict downstream policy learning. As shown in Table~\ref{tab:value_ablation}, the reliability ranking aligns with downstream success: \textsc{Short History} provides the strongest overall diagnostic performance, and also achieves the highest success rates on both precise chip insertion and generalizable block disassembly tasks. 
This alignment indicates that the metrics capture value-estimation properties that are relevant to policy optimization.
This suggests that value functions capturing progress consistency, local smoothness, and error discrimination better guide policy optimization, making our metrics a lightweight criterion for value-estimator selection before expensive downstream training.

\subsection{Value-Guided Offline Policy Pretraining}
\label{sec:offline_policy_learning}

Offline policy pretraining on heterogeneous robotic data hinges on selecting useful experience from demonstrations, corrections, suboptimal actions, and failures. We study how different value estimators and action-quality selection strategies shape policy learning from such mixed-quality data.

\textbf{Scaling with Heterogeneous Offline Data.} We evaluate whether value-guided pretraining can better exploit heterogeneous offline data by training chip-insertion and block-disassembly policies on mixed-quality subsets of 10, 50, and 120 hours. We compare four variants: Behavior Cloning (\textsc{BC}) uses all data without value-based quality labels, treating every demonstrated action equally regardless of its contribution to task progress; Value-Guided Soft (\textsc{VG-Soft}) derives action-quality labels from the \textsc{Short History} value estimator using a longer temporal window ($\Delta{=}60$ frames) and relaxed positive margins to select the top 50\% of samples, which retains more diverse behaviors including temporally extended manipulation sequences; Value-Guided Strict (\textsc{VG-Strict}) uses the same value estimator but with a shorter temporal window ($\Delta{=}20$ frames) and tighter margins to select the top 30\% of samples, emphasizing immediate action-level progress; and Value-Guided No History (\textsc{VG-NH}) applies the same strict labeling rule as \textsc{VG-Strict} but uses the \textsc{No History} value estimator, isolating the effect of value-estimation reliability on downstream policy learning.

\begin{figure*}[t]
    \centering
\includegraphics[width=\columnwidth]{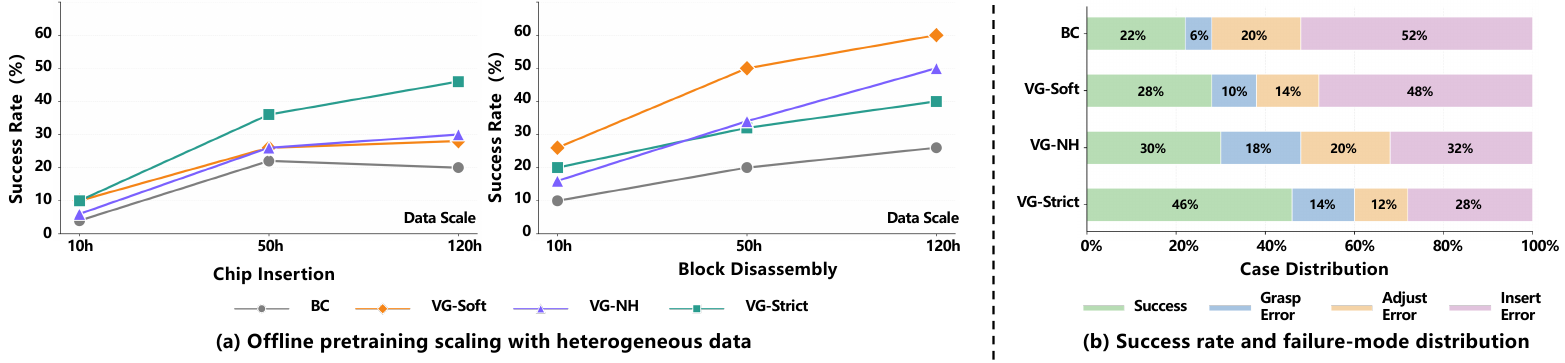}
    \caption{Value-guided offline pretraining performance and behavior. (a) \textsc{BC} saturates as mixed-quality data increases, while value-guided variants scale more effectively. (b) Value-estimation reliability directly shapes downstream policy behavior.}
    \label{fig:scaling_offline}
    \vspace{-1em}
\end{figure*}

As shown in Figure~\ref{fig:scaling_offline}(a), \textsc{BC} yields limited gains beyond 50 hours, since \textsc{BC} treats all samples uniformly; additional data may introduce low-quality supervision from suboptimal or failed behaviors. In contrast, value-guided pretraining emphasizes high-value samples, enabling the policy to learn high-quality action distributions and scale effectively with heterogeneous offline data.
These results in Figure~\ref{fig:scaling_offline}(b) show that quality threshold is task-dependent. \textsc{VG-Strict} performs better on chip insertion, where progress depends on short precise motions, whereas \textsc{VG-Soft} is better suited to block disassembly, where progress spans longer manipulation segments and requires broader behavioral coverage.
The comparison between \textsc{VG-Strict} and \textsc{VG-NH} further connects these results to our value-reliability metrics in Sec.~\ref{sec:value_diagnostics}. Both methods use the same strict labeling rule, but differ in the value estimator used to assign action-quality labels. The stronger performance of \textsc{VG-Strict} indicates that the more reliable \textsc{Short History} estimator produces more accurate quality labels than the \textsc{No History} estimator. This shows that value-estimation reliability directly affects downstream offline reinforcement learning by determining which actions are selected as beneficial training signals.

\textbf{Failure Case Analysis.}
We further analyze how value-based quality criteria reshape policy behavior. BC assumes all demonstrated actions are valid, thereby preserving routine behaviors well: because human demonstrations rarely fail during grasping, BC yields the lowest grasp-error rate. However, it also imitates hesitation and trial-and-error corrections during insertion, leading to frequent insertion failures. Value-guided pretraining shifts this behavior distribution by selecting high-quality action chunks. Stricter filtering suppresses noisy insertion-stage corrections, reducing insertion errors and improving success, but may also remove supervision for routine stages such as grasping and coarse adjustment, increasing early-stage failures. The gap between \textsc{VG-Strict} and \textsc{VG-NH} further shows that this trade-off depends on value reliability: unreliable estimates can discard useful chunks and amplify early failures.

\subsection{Value-Guided Online Improvement}
\label{sec:online_adaptation}
Online rollouts help bridge the distribution gap between offline pretraining data and real-world interaction, but their heterogeneous nature makes them difficult to exploit effectively. We therefore study how value guidance can stabilize online policy optimization by identifying high-quality interaction data and suppressing suboptimal updates.

\begin{wrapfigure}{r}{0.42\linewidth}
\vspace{-0.8em}
\centering
\includegraphics[
width=\linewidth,
]{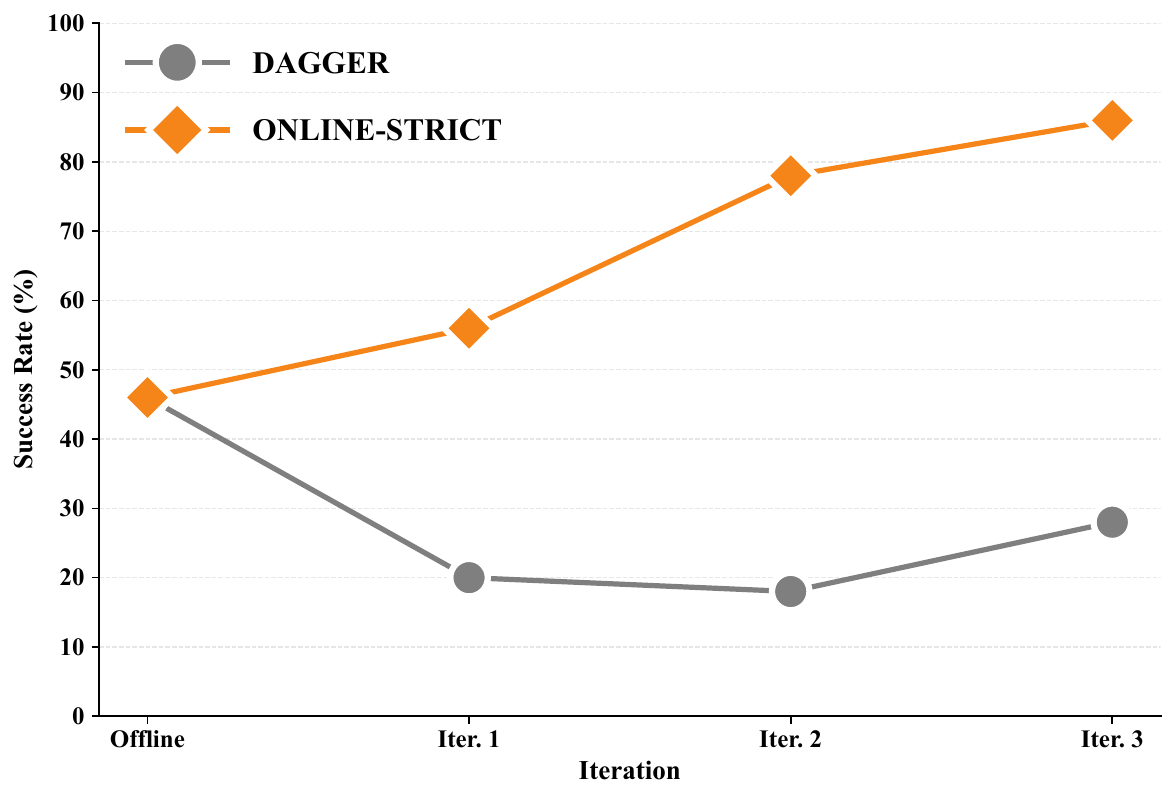}
\vspace{-0.8em}
\caption{
Iterative online improvement on the chip insertion task.
}
\label{fig:online_iteration}
\vspace{-1.2em}
\end{wrapfigure}

\textbf{Multi-Iteration Online Improvement.}
Starting from the \textsc{VG-Strict} offline-RL-pretrained policy, which achieves a 46\% success rate (SR) on the chip insertion task, we conduct three rounds of online rollout collection and policy improvement, using 500 rollout trajectories per iteration. We compare against Dataset Aggregation (\textsc{DAgger})~\cite{ross2011reductionimitationlearningstructured}, a widely used online imitation learning method that aggregates human-corrected trajectories into the training set and retrains the policy; here it directly treats all human corrections as positive supervision without distinguishing action quality. In contrast, Value-Guided Online Strict (\textsc{Online-Strict}) initializes the value model from \textsc{VG-Strict} and further fine-tunes it during each online iteration using the same strict value-estimation protocol.
The updated value model is then used to score online rollout segments and select high-quality actions for policy improvement.

As shown in Figure~\ref{fig:online_iteration}, \textsc{Online-Strict} improves from 46\% to 86\% over three online iterations. This suggests that the iteratively fine-tuned value model can identify useful correction signals from heterogeneous rollouts and convert them into stable policy improvement. 
In contrast, \textsc{DAgger} drops sharply, and its adapted performance approaches the offline \textsc{BC} baseline. This suggests that aggregating corrective trajectories without quality assessment harms online improvement due to the suboptimal behaviors present in human correction data.

\begin{wraptable}{r}{0.3\linewidth}
\vspace{-0.8em}
\centering
\caption{
Online policy improvement under different value estimation settings after one iteration of improvement.
}
\label{tab:online_ablation_wrap}
\vspace{-0.6em}
\setlength{\tabcolsep}{4pt}
\renewcommand{\arraystretch}{0.95}
\resizebox{\linewidth}{!}{
\begin{tabular}{lcc}
\toprule
\textbf{Method} & \textbf{Chip} & \textbf{Block} \\
\midrule
\textsc{DAgger} & 20\% & 66\% \\
\textsc{online-Soft} & 32\% & \textbf{84\%} \\
\textsc{online-Strict}  & \textbf{56\%} & 70\% \\
\textsc{online-NH}       & 40\% & 70\% \\
\bottomrule
\end{tabular}
}
\vspace{-1.2em}
\end{wraptable}

Value-Guided Online Soft (\textsc{Online-Soft}) and Value-Guided Online No History (\textsc{Online-NH}) follow the same online pipeline as \textsc{Online-Strict}, but initialize the value model from \textsc{VG-Soft} and \textsc{VG-NH}, respectively. As shown in Table~\ref{tab:online_ablation_wrap}, all value-guided online variants outperform \textsc{DAgger}, highlighting the benefit of value-based quality filtering for online rollout data. The performance ranking remains task-dependent: \textsc{Online-Strict} achieves the best success rate on precision-critical chip insertion, while \textsc{Online-Soft} performs best on block disassembly, which requires broader temporal coverage to preserve diverse manipulation sequences. The gap between \textsc{Online-Strict} and \textsc{Online-NH} further demonstrates that value-estimation reliability directly impacts online improvement quality.

\subsection{Qualitative Analysis}
\label{sec:qualitative}

Beyond the quantitative results presented above, we provide qualitative analysis to demonstrate the effectiveness of our Robo-ValueRL framework. We visualize the gate activation patterns of the online residual adapter and present representative behavioral patterns acquired by the robot, including both optimal behaviors and self-correction capabilities.

\begin{figure*}[h]
    \centering
\includegraphics[width=\columnwidth]{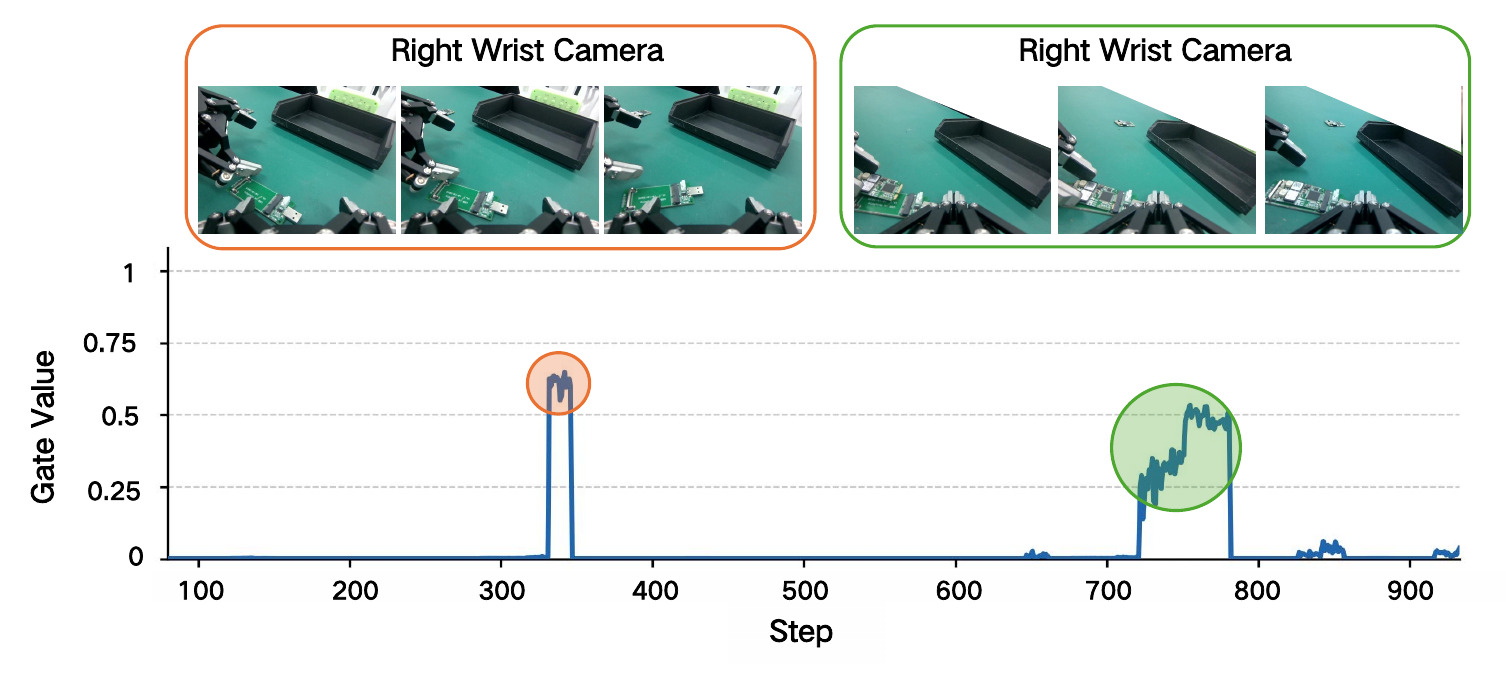}
\vspace{-0.5em}
    \caption{Gate activation pattern of the online residual adapter over time. The online module mostly remains inactive when the offline policy is sufficient. It is selectively triggered at critical error-prone stages, such as PCB pose adjustment (left) and chip insertion (right), to provide targeted action refinement.}
    \label{fig:gate_main}
\end{figure*}

\textbf{Gate Activation of Online Residual Adapter.}
We visualize the gate activation pattern of the online residual adapter during task execution in Figure~\ref{fig:gate_main}. For most steps, the offline policy already performs reliably, so the gate remains inactive and the residual adapter does not intervene. At critical error-prone states, such as PCB orientation adjustment and fine-grained chip insertion, the gate is activated to enable targeted refinement. The left panel shows correction of erroneous PCB adjustment, while the right panel shows fine-grained action refinement during precision insertion. This selective activation indicates that the gate learns to identify when the base policy is insufficient, preserving the pretrained behavior prior while enabling precise online correction.


\begin{figure*}[t]
\centering
\includegraphics[
width=\textwidth,
keepaspectratio
]{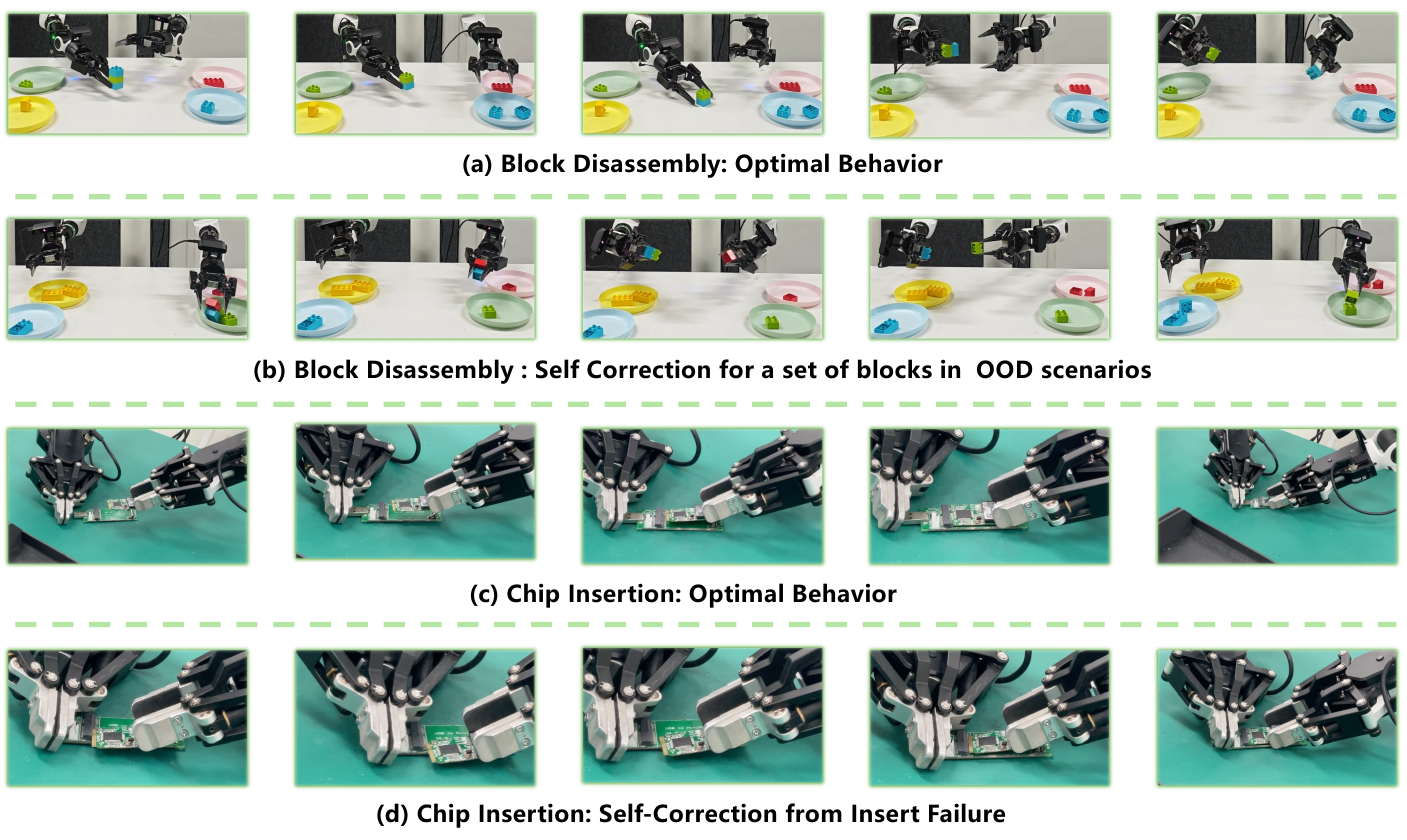}
\vspace{-2em}
\caption{
Representative learned behavioral patterns in block disassembly and chip insertion.
}
\label{fig:pattern_main_all}
\vspace{-1em}
\end{figure*}

\textbf{Learned Behavioral Patterns.}
Figure~\ref{fig:pattern_main_all} presents representative behaviors in block disassembly and chip insertion, highlighting two capabilities learned through our offline-to-online reinforcement learning framework: efficient optimal execution and autonomous self-correction. Additional patterns are provided in Appendix~\ref{sec:pattern}.


\textit{Optimal Behavior.}
As shown in Figure~\ref{fig:pattern_main_all}(a), the robot learns non-greedy disassembly strategies that anticipate future placement constraints. Although immediately disassembling the next block may appear locally optimal, it would require extra tabletop placement and hand-switching before final placement. Instead, the robot first releases the blue block, then assigns the green and blue blocks to the hands aligned with their target trays, enabling direct placement after disassembly. This shows that value-guided reinforcement learning favors long-horizon efficiency over greedy local progress. In chip insertion (Figure~\ref{fig:pattern_main_all}(c)), the robot further completes millimeter-level insertion in a single attempt, whereas human teleoperators typically require 2--3 attempts, yielding over 1.5$\times$ higher execution efficiency.


\textit{Self-Correction.}
As shown in Figure~\ref{fig:pattern_main_all}(b), when a block falls to an unexpected location, the robot autonomously retrieves it and returns it to the correct tray. In chip insertion (Figure~\ref{fig:pattern_main_all}(d)), when an insertion failure occurs, the robot lifts the gripper and performs a second insertion attempt. These recovery behaviors emerge from online improvement with value-filtered rollout data, where human-corrected trajectories teach the policy to associate failure states with corrective actions, enabling autonomous error recovery during deployment.

\section{Conclusion}

We presented Robo-ValueRL, an offline-to-online reinforcement learning framework that enables scalable offline pretraining and stable online improvement through reliable value estimation. Robo-ValueRL learns a history-conditioned value estimator, propagates its predictions into quality-conditioned consistency-policy pretraining, and uses online rollouts to train a lightweight residual adapter for targeted adaptation while preserving the pretrained prior. We assess value reliability from global-progress and local-preference perspectives, showing that reliability strongly predicts downstream performance: more reliable values lead to better offline scaling and more stable online improvement. By integrating value guidance across offline pretraining and online adaptation, Robo-ValueRL achieves 86\% success on millimeter-level chip insertion and 84\% on generalizable block disassembly. We hope this study encourages future robotic learning systems to move beyond simple data accumulation and toward reliable value-guided data utilization.


\clearpage

\section{Acknowledgement}

This work is supported by Beijing Natural Science Foundation (4262050). This work is also supported by Beijing Innovation Center of Humanoid Robotics (X-Humanoid).
\bibliographystyle{IEEEtran}
\bibliography{paper}

\clearpage

\beginappendix

\section{Overview}

In this work, we investigate \textit{how value-function reliability affects value-guided learning across offline pretraining and online improvement}. To answer this question, we build a complete offline-to-online reinforcement learning pipeline. In the supplementary materials, we first present the performance of the model trained through the complete pipeline on two representative real-world tasks: a millimeter-level precise chip-insertion task and a generalizable block-disassembly task. In Section~\ref{sec:one_task} and the supplementary video, we provide \textbf{long-duration, one-take real-world videos of over 35 minutes for each task}. The block-disassembly task achieves an \textbf{82.9\%} success rate \textbf{(58/70)}, while the chip-insertion task achieves an \textbf{85.7\%} success rate \textbf{(30/35)}.
We also provide detailed implementation details to support reproducibility in Section~\ref{sec:details}.
Beyond these quantitative results, in Section~\ref{sec:pattern}, we further demonstrate several surprising generalization and self-correction capabilities acquired by the robot through our offline-to-online framework. These include recovering from mistakes, completing tasks in out-of-distribution scenarios, and exhibiting more efficient action patterns than skilled human demonstrators. Next, in Section~\ref{sec:value_sup}, we provide visualizations of the value estimator and the corresponding action quality, showing that our framework can effectively estimate task progress in manipulation tasks and further predict action quality to guide policy learning. 

\section{One-take Real-world Experiment Video}\label{sec:one_task}
We introduce the task setting and provide one-take real-world experiment videos in the supplementary material. As shown in Figure~\ref{fig:task}, we evaluate our system on two tasks: a millimeter-level chip-insertion task, where the robot must precisely insert a chip into a narrow slot, and a generalizable block-disassembly task, where the robot disassembles randomly placed blocks and sorts them into color-matched plates under randomized layouts.

The supplementary video includes continuous one-take real-world experiments for both tasks. For block disassembly, the video is played at 15$\times$ speed, and the robot achieves an 82.9\% success rate (58/70) over a continuous 35-minute experiment. For chip insertion, the video is played at 10$\times$ speed, and the robot achieves an 85.7\% success rate (30/35) over a continuous 40-minute experiment. Several frame-drop events occurred during chip insertion due to instability of the Orbbec Gemini 336 camera SDK, causing intermittent pauses and slightly affecting the final success rate.




\section{Implementation Details}\label{sec:details}

\subsection{Value Estimator}\label{sec:value_estimator_sup}
We build the value estimator using a PaliGemma VLM backbone~\cite{beyer2024paligemma} and a transformer-based value head. To leverage the knowledge learned from large-scale robotic datasets, we initialize the PaliGemma backbone from the $\pi_0$~\cite{black2024pi_0} VLM checkpoint and keep it frozen during training. We mix the OpenX dataset~\cite{o2024open}, our self-collected data, the chip-insertion and block-disassembly datasets, which are important for preventing overfitting since the supervised value-prediction objective can otherwise be easily overfit. We train the value estimator for two epochs with an initial learning rate of $1\times10^{-4}$, which is decayed to $5\times10^{-6}$ using a cosine scheduler.
During online improvement, we collect rollout data and mix it with the offline datasets at a fixed ratio of 3:1, followed by another two epochs of value-estimator training.

For value estimation, we use a discrete distributional head and formulate value prediction as classification over $K=256$ fixed bins. We first compute a remaining-time target for each timestep $t$ in trajectory $\tau$. For successful trajectories, the remaining time is the number of steps before task completion. For failed trajectories, we add a failure penalty $C_{\mathrm{fail}}$ to assign lower values to all timesteps:
\begin{equation}
r_t^* =
\begin{cases}
T_\tau - t, & \text{if } \tau \text{ is successful}, \\
T_\tau - t + C_{\mathrm{fail}}, & \text{otherwise},
\end{cases}
\end{equation}
where $T_\tau$ denotes the trajectory length. We then normalize the remaining-time target into a progress-style value target in $[0,1]$:
\begin{equation}
v_t^* = \mathrm{clip}\left(1 - \frac{r_t^*}{T_{\max}}, 0, 1\right),
\end{equation}
where $T_{\max}=5000$. This formulation assigns larger values to states closer to successful completion, while failed trajectories receive lower values due to the additional penalty.

To train the distributional value head, we discretize $[0,1]$ into $K$ equally spaced bins with centers $c_k = k/(K-1)$ for $k=0,\dots,K-1$. Instead of using a hard one-hot label, we convert $v_t^*$ into a soft target distribution:
\begin{equation}
p_k(v_t^*) =
\frac{
\exp\left(-\frac{(v_t^* - c_k)^2}{2\sigma^2}\right)
}{
\sum_{j=0}^{K-1}
\exp\left(-\frac{(v_t^* - c_j)^2}{2\sigma^2}\right)
},
\quad
\sigma = \frac{1}{K-1}.
\end{equation}
The value estimator is trained by minimizing the cross-entropy between the predicted distribution and this soft target distribution.

\subsection{Quality-conditioned Consistency Policy}\label{sec:policy_sup}

To convert the action-quality indicator $q_t^{\mathrm{act}}$ into a textual action-quality prompt $\ell_t^{\mathrm{act}}$, we use a rule-based mapping:
\begin{equation}
\ell_t^{\mathrm{act}} =
\begin{cases}
\texttt{Quality: Low}, & q_t^{\mathrm{act}} = 0,\\
\texttt{Quality: Medium}, & q_t^{\mathrm{act}} = 1 \\ 
\texttt{Quality: High}, & q_t^{\mathrm{act}} = 2.
\end{cases}
\end{equation}
During training, we further apply quality-condition dropout: with a probability of 10\%, we replace $\ell_t^{\mathrm{act}}$ with \texttt{Medium}, regardless of the original quality label.

For the action expert, we instantiate $f_\theta^a$ as a consistency policy, where the noise level $\sigma$ is sampled from $[2,80]$ during training. To improve training stability, we adopt a noise-level curriculum that starts with low-noise samples and gradually expands to higher noise levels. The full VLA model $\pi_{\theta}$ is trained for 50,000 steps on 32 A100 GPUs, with the learning rate decayed from $2\times10^{-4}$ to $5\times10^{-6}$ with a cosine scheduler.

\subsection{Online Adapter}\label{sec:adapter_module_sup}
During online improvement, we freeze the offline policy and train only an online adapter. The adapter is zero-initialized so that, at the beginning of online training, it behaves as a near-identity correction and does not significantly deviate from the offline policy. In each update round, we mix online and offline data with a 3:1 ratio and limit training to 900 gradient steps, preventing the gate from overfitting to the limited online rollouts. 

During online interaction, we find that the robot often fails by getting stuck. For example, in chip insertion, the chip may get stuck on the PCB base plate and fail to lift, while in block disassembly, the gripper may push against the inner block and fail to grasp it. These failure modes make the representations of human-corrected states highly similar to those of prolonged stuck states. Therefore, we use high-quality samples from the online interaction dataset $\mathcal{D}_{\mathrm{on}}$ to train the gate to open, while using only offline data to train the gate to remain closed. This allows the model to acquire corrective behaviors for previous low-quality states through representation similarity, while avoiding the ill-posed supervision problem of learning corrections from low-quality samples without ground-truth corrected actions.


\section{Learned Pattern from Our Methods}~\label{sec:pattern}

\begin{figure}[H]
    \centering
    \includegraphics[
        width=\textwidth,
        keepaspectratio
    ]{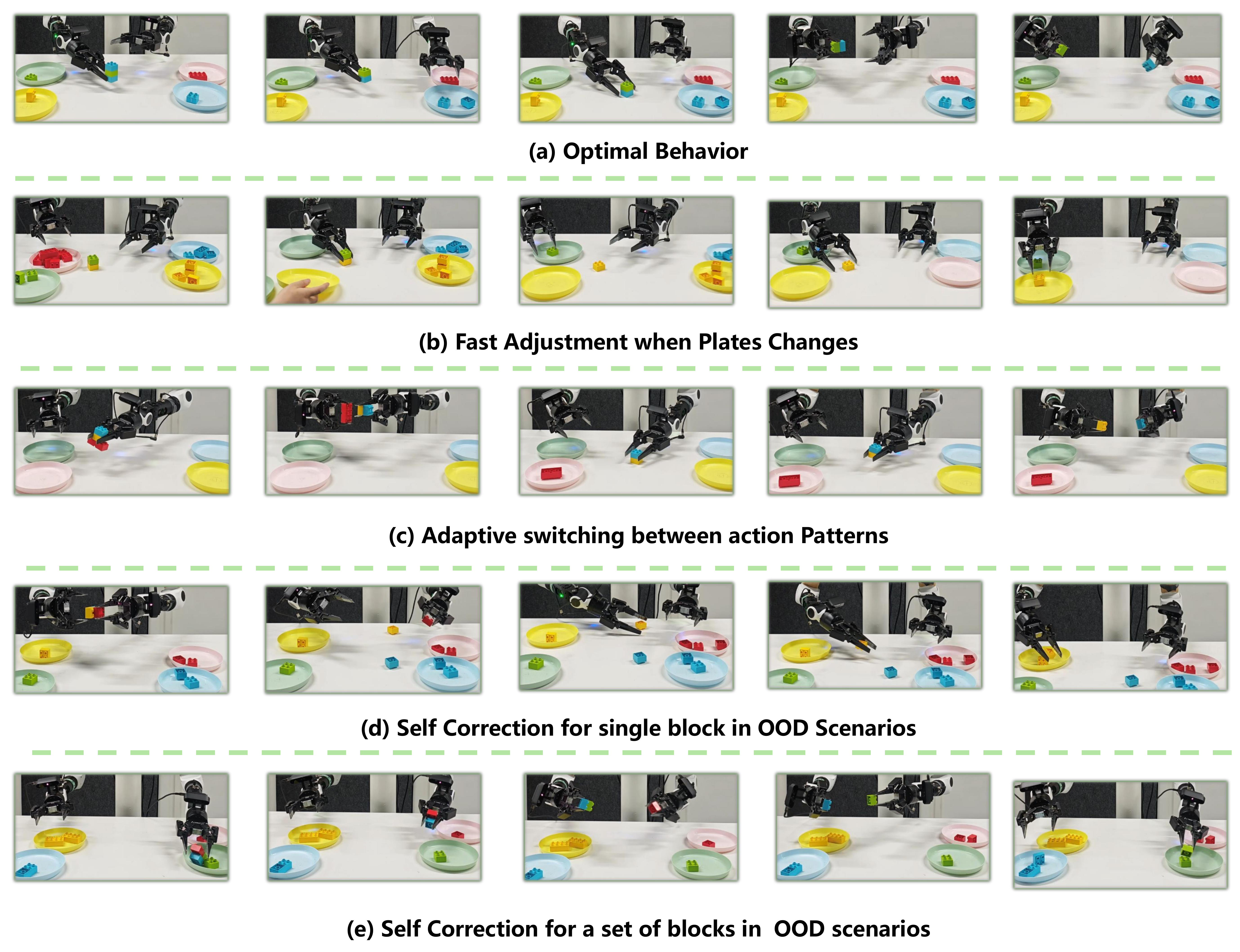}
    \caption{
    We demonstrate several robot self-correction and optimal-action segments in the block-disassembly task.
    These include autonomously selecting the most appropriate disassembly strategy to maximize efficiency,
    rapidly responding to scene changes, and self-correcting under out-of-distribution conditions.
    }
    \label{fig:Pattern}
\end{figure}

\begin{figure}[H]
    \centering
    \includegraphics[
        width=\textwidth,
        keepaspectratio
    ]{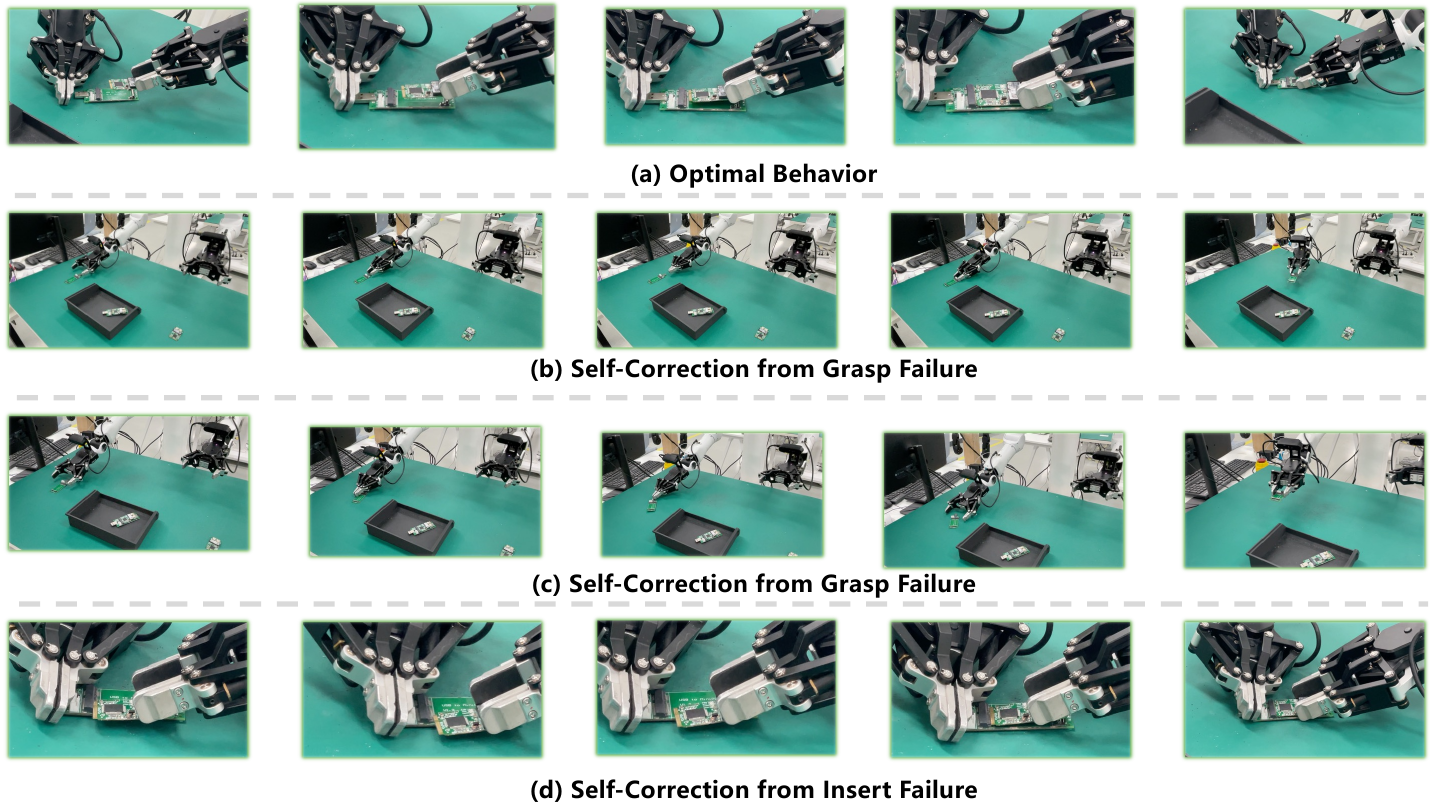}
    \caption{
    We demonstrate robot self-correction and optimal-action segments in the chip-insertion task.
    The robot adapts its manipulation strategy under execution disturbances and recovers from unstable intermediate states.
    }
    \label{fig:Pattern_chip}
\end{figure}

We present the representative behavioral patterns in Figure~\ref{fig:Pattern} and Figure~\ref{fig:Pattern_chip}, we also provide the corresponding videos to demonstrate our robots' behaviors.

For the block-disassembly task, we present five representative behavioral patterns exhibited by our agent in Figure~\ref{fig:Pattern}. As shown in (a), the robot learns optimal disassembly sequences, completing the task efficiently without redundant actions. After the robot finishes disassembling the previous block, its left hand is holding the blue block. If it proceeds with disassembly at this point, the left hand will still hold the blue block while the right hand will hold the green block. However, since the blue tray is on the right and the green tray is on the left, the blocks cannot be placed directly into their corresponding plates.
\textbf{Instead, the robot chooses to put down the block first.} It then uses its left hand to grasp the upper green block and its right hand to disassemble the upper blue block, so that after disassembly, each block can be directly placed into the tray with the matching color. By using task progress as guidance for RL learning, the robot learns optimal actions that maximize task-completion efficiency.
In (b), when the target plate is suddenly moved to the opposite side during reaching, the agent immediately detects the change, retracts its extended hand, and switches to the other hand to complete the placement. In (c),
The robot grasps the middle of the three-block assembly and first disassembles the bottom red block by pulling it horizontally. However, the same action pattern can no longer continue the disassembly. Therefore, the robot switches to a different manipulation mode, reorients the blocks upright, and then disassembles another block. This avoids unnecessary hand swapping.
In (d), when a single block falls to a distant out-of-distribution location during disassembly, the agent recognizes the error and retrieves the block back to the correct plate. In (e), when a whole set of blocks is placed into the plate without being disassembled first, the agent detects the mistake, picks them up again, and performs the correct disassembly sequence. These patterns collectively demonstrate the agent's robustness, adaptability, and strong generalization ability

For the chip-insertion task, we demonstrate diverse robot behaviors in Figure~\ref{fig:Pattern_chip}. As shown in (a), our robot can perform chip insertion highly efficiently. Unlike human teleoperators, who typically require 2--3 attempts, our robot can complete precise millimeter-level insertion in a single attempt, achieving more than 1.5$\times$ the efficiency of human teleoperators. (b) and (c) show that when grasping failures occur, the robot can autonomously correct its behavior and re-grasp the chip. (d) shows that when an error occurs, the robot can lift the gripper and perform a second insertion attempt.

\section{Value Estimator and Action Quality Estimation Performance}\label{sec:value_sup}

\begin{figure}[H]
    \centering
    \includegraphics[
        width=0.85\textwidth,
        keepaspectratio
    ]{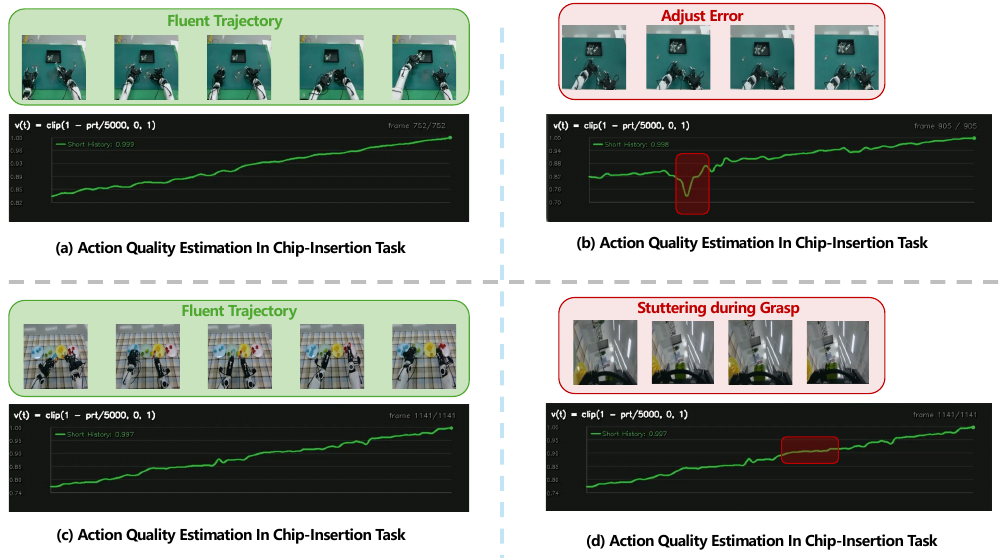}
    \vspace{-0.5em}
    \caption{\textbf{Value-estimation results.} We visualize value estimates on chip insertion and block manipulation tasks. Successful trajectories produce smooth and stable value curves, while execution errors or stagnation are reflected by sharp value drops or low-amplitude fluctuations, showing that the value estimator captures task progress and failure recovery.}
    \label{fig:value_estimation}
\end{figure}

\begin{figure}[H]
    \centering
    \includegraphics[
        width=0.85\textwidth,
        keepaspectratio
    ]{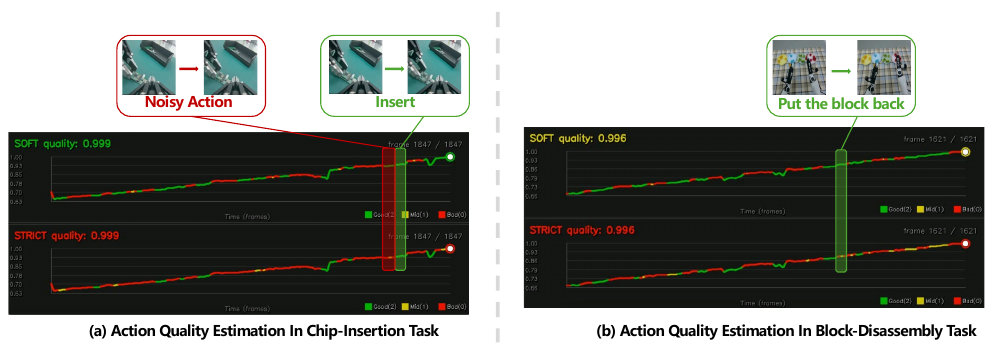}
    \vspace{-0.5em}
    \caption{\textbf{Effect of action-quality filtering.} We compare strict and soft action-quality thresholds under the same value estimates. Strict filtering better identifies effective short-horizon actions in the precise chip-insertion task, whereas soft filtering is more suitable for the long-horizon block task by preserving more useful but temporally extended actions.}
    \label{fig:action_quality}
\end{figure}

We present the results of value estimation and analyze how action quality affects action filtering with the corresponding video. In Figure~\ref{fig:value_estimation}, we show the value-function estimation results on the chip-insertion task and the block-sorting task. As shown in Figures~\ref{fig:value_estimation}(a) and (c), when a trajectory can smoothly complete the task, the estimated value curve remains stable and smooth throughout the execution. In contrast, when errors occur, our model can sensitively capture their impact on task progress and reflect this change in the value curve. For example, in Figure~\ref{fig:value_estimation}(b), when an adjustment error occurs and the left hand fails to properly align the PCB base to the desired pose, the estimated value drops sharply. It then recovers as the base is manually corrected to the proper position. Figure~\ref{fig:value_estimation}(d) further shows that when the model remains stuck in the grasping stage for an extended period, the value function only fluctuates within a small range, indicating that the task progress is not being advanced.
We further visualize the effect of different action-quality thresholds under the same value-function estimates in Figure~\ref{fig:action_quality}. Specifically, the strict criterion selects the top 30\% of samples based on $v(t+20)-v(t)$, whereas the soft criterion selects the top 50\% of samples based on $v(t+60)-v(t)$. As shown in the figure, chip insertion is a highly precise manipulation task, where even a short action segment of around 10 frames can determine whether the insertion succeeds. Therefore, the strict threshold can effectively identify meaningful and efficient actions. In contrast, the soft threshold tends to label many suboptimal pre-insertion actions as good actions, which can negatively affect policy learning. The block task exhibits the opposite behavior. Its disassembly process consists of long-horizon actions that require stronger generalization but lower precision. In this case, an overly strict threshold may incorrectly filter out many useful actions as bad ones, thereby degrading the learned policy.

\end{document}